\documentclass[10pt]{article}

\usepackage{amsmath}
\usepackage{placeins}
\usepackage{placeins}
\usepackage{amssymb}
\usepackage{amsthm}
\usepackage{booktabs}       

\usepackage{graphicx}

\usepackage{cite}
\usepackage{color} 
\usepackage[colorlinks,breaklinks]{hyperref}
\hypersetup{hypertexnames=true,
  linkcolor=blue,anchorcolor=black,citecolor=blue,urlcolor=blue
}

\DeclareGraphicsExtensions{.png,.pdf}
\graphicspath{{imglocal/}{img/}{./}} 

\usepackage[labelfont=bf,labelsep=period,justification=raggedright]{caption}
\usepackage{subcaption}

\makeatletter
\renewcommand{\@biblabel}[1]{\quad#1.}
\makeatother

\newcommand{\giacomo}[1]{}

\date{}

\begin{document}

\begin{flushleft}
{\Large
\textbf{Deep supervised learning using local errors}
}
\\
Hesham Mostafa$^1$, 
Vishwajith Ramesh$^2$, and
Gert Cauwenberghs$^{1,2}$
\\
$^{1}$Institute for Neural Computation,$^{2}$Department of Bioengineering \\
UC San Diego, California, USA \\
E-mail: hmmostafa@ucsd.edu
\end{flushleft}

\section*{Abstract}
Error backpropagation is a highly effective mechanism for learning high-quality hierarchical features in deep networks. Updating the features or weights in one layer, however, requires waiting for the propagation of error signals from higher layers. Learning using delayed and non-local errors makes it hard to reconcile backpropagation with the learning mechanisms observed in biological neural networks as it requires the neurons to maintain a memory of the input long enough until the higher-layer errors arrive. In this paper, we propose an alternative learning mechanism where errors are generated locally in each layer using fixed, random auxiliary classifiers. Lower layers could thus be trained independently of higher layers and training could either proceed layer by layer, or simultaneously in all layers using local error information. We address biological plausibility concerns such as weight symmetry requirements and show that the proposed learning mechanism based on fixed, broad, and random tuning of each neuron to the classification categories outperforms the biologically-motivated feedback alignment learning technique on the MNIST, CIFAR10, and SVHN datasets, approaching the performance of standard backpropagation. Our approach highlights a potential biological mechanism for the supervised, or task-dependent, learning of feature hierarchies. In addition, we show that it is well suited for learning deep networks in custom hardware where it can drastically reduce memory traffic and data communication overheads.


\section{Introduction}
Gradient descent training techniques~\cite{Bottou91} have been remarkably successful in training a broad range of network architectures. This success is often attributed to the use of deep architectures with many non-linearity stages~\cite{Ba_Caruna14} where backpropagation is used to calculate the direction of weight updates in deep layers. In convolutional networks in particular, multiple cascaded convolutional layers allow simple, lower-level, features to be successively composed into more complex features, allowing networks to obtain highly complex and relevant features from the top convolutional layers~\cite{Razavian_etal14}. Deep convolutional neural networks trained using backpropagation are thus achieving record performance in a variety of large-scale machine vision tasks~\cite{Krizhevsky_etal12,Simonyan_Zisserman14,LeCun_etal15,He_etal16,Zagoruyko_Komodakis16,Huang_etal16}. For deep convolutional networks trained in a supervised setting, the training objective is typically the minimization of classification error at the top network layer. This objective is sometimes augmented by auxiliary objectives defined using the outputs of intermediate classifiers in the network~\cite{Szegedy_etal14,Lee_etal15}. These auxiliary objectives provide additional sources of error to deeper layers. Training, however,  involves error signals that must propagate backwards from the top layer. 

Standard backpropagation is biologically unrealistic for several reasons: the need to buffer network states until errors arrive from the top layer; weight symmetry in the forward and backward passes; and the need to precisely interleave the forward and backward passes. Several biologically-motivated learning mechanisms have been proposed to explain how circuits in the brain are able to learn complex, hierarchical representations. One broad class of these proposals is based on contrastive learning in energy-based models~\cite{Seung_etal03,Bengio_Fischer15,Scellier_Bengio17}. In these models, the network is trained to minimize the discrepancy between its equilibrium points when running freely and when observables clamp the values of some units in the network. Weight symmetry is required, though: each synaptic connection from one neuron to another assumes a matching synaptic connection of identical strength in the reverse direction. In~\cite{Lillicrap_etal16,Baldi_etal16}, weight symmetry is avoided by using an independent set of fixed random weights to backpropagate errors between the network layers. However, like standard backpropagation, the error signals are non-local. Instead of backpropagating errors layer by layer through the random feedback connections, the networks in~\cite{Nokland_etal16,Neftci_etal17a} directly use a fixed random projection of the top layer error as the error signal in deep layers. Although this permits a single global error signal communicated in common to all layers, is still incurs substantial wait times and memory requirements for the weight updates as a forward pass through the entire network has to be completed before the error signal is available, which requires deep layers to hold their states for the duration of the full forward pass.

We propose a learning approach where weights in any given layer are trained based on local errors that are generated solely based on neural state variables in that layer. These errors are generated directly from the training labels using a classifier with fixed random weights and no hidden layers, and whose input is the neural activations in the layer being trained. Instead of minimizing a global objective function, training thus minimizes many local objective functions. As such this approach compromises one of the core tenets of standard backpropagation: the adjustment of all parameters in concert to minimize a unified objective.  Nevertheless, training with local errors still allows a deep network to compose the features learned by lower layers into more complex features in higher layers. This is evidenced below by the improvement in accuracy of the random local classifiers in deeper layers. Training with local errors thus retains the hierarchical composition of features, one of the key strengths of deep networks. 

To implement weight updates based on backpropagation in a biologically inspired network, the pre- or post-synaptic neurons need to buffer the past activity of the pre-synaptic neurons and reproduce this past activity in sync with the corresponding errors arriving from top layers in order to update the weights. This is incompatible with biologically motivated synaptic weight update rules that are typically triggered by pre-synaptic events and depend on the relative timing of pre- and post-synaptic spikes and/or state variables in the post-synaptic neuron. Our learning mechanism bypasses biological implausibility arguments against standard backpropagation by generating errors locally in each layer using fixed random projections.
Weight updates could thus be carried out while the synaptic currents in post-synaptic neurons (the neurons receiving the local error signal) still retain a memory of recent pre-synaptic activity. Weight symmetry in the forward and backward passes in standard backpropagation learning is another biologically unrealistic aspect. In our case, the weight symmetry requirement arises in the one-step error backpropagation from the output of the local random classifier to the neurons in the layer being trained. Similar to ref.~\cite{Lillicrap_etal16}, we experimented with relaxing this symmetry requirement by using a different set of random, fixed weights to map the classifier error to the error at the layer being trained. 

We analyze the implications of the proposed learning approach for the design of custom hardware devices for learning the parameters of deep networks. In the proposed learning approach, there is no explicit backward pass as errors are locally generated and can be used to directly update the weights. We show that our approach  drastically reduces memory traffic compared to standard backpropagation in the typical situation when the network weights and activations can not all fit into the compute device memory. We achieve this reduction even despite an increased number of parameters in the network due to the addition of the random local classifier weights in each layer. These weights, however, are fixed allowing them to be generated on the fly using pseudo-random number generators (PRNGs). Only the negligibly small random seeds of the PRNGs for each layer need to be stored. 

We discuss related work in section~\ref{sec:related_work}. We describe the proposed learning mechanism in section~\ref{sec:model} and quantitatively assess the hardware-related computational and memory access benefits compared to standard learning with global objective functions in section~\ref{sec:hw}. We present the results of applying the proposed learning method to standard supervised learning benchmarks in section~\ref{sec:results} and compare our learning method's performance to that of the feedback alignment technique~\cite{Lillicrap_etal16} . We present our conclusions and further discussion on the biological plausibility of the proposed learning mechanism in section~\ref{sec:conclusions_discussion}.

\section{Related Work}
\label{sec:related_work}
Training of deep convolutional networks is currently dominated by approaches where all weights are simultaneously trained to minimize a global objective. This is typically done in a purely supervised setting where the training objective is the classification loss at the top layer. To ameliorate the problem of exploding/vanishing errors in deep layers~\cite{Hochreiter_etal01}, auxiliary classifiers are sometimes added to provide additional error information to deep layers~\cite{Szegedy_etal14,Lee_etal15}. Unlike our training approach, however, training still involves backpropagating errors across the entire network and simultaneous adjustments of all weights.

Several learning mechanisms have been traditionally used to pre-train a deep network layer-by-layer using local error signals in order to learn the probability distribution of the input layer activations, or in order to  minimize local reconstruction errors~\cite{Hinton_etal06,Hinton_Salakhutdinov06,Bengio_etal07,Vincent_etal08,Erhan_etal10}. These mechanisms, however, are unsupervised and the networks need to be augmented by a classifier layer, typically added on top of the deepest layer. The network weights are then fine-tuned using standard backpropagation to minimize the error at the classifier layer.  Supervised layer-wise training has been pursued in \cite{Bengio_etal07}, with auxiliary classifiers that are co-trained, unlike the random fixed auxiliary classifiers proposed here. 
The supervised layer-wise training is used only as a pre-training step, and results are reported after full network fine-tuning using backpropagation from the top classifier layer. Some approaches forego the fine-tuning step and keep the network fixed after the unsupervised layer-wise training phase, and only train the top classifier layer or SVM on the features learned~\cite{Lee_etal09,Ranzato_etal07,Kavukcuoglu_etal10}. Local learning in~\cite{Ranzato_etal07,Kavukcuoglu_etal10} involves an iterative procedure for learning sparse codes which is computationally demanding. 
The network architectures in \cite{Lee_etal09,Ranzato_etal07,Kavukcuoglu_etal10} fail to yield intermediate classification results from the intermediate layers. Moreover, their applicability to datasets that are more complex than MNIST is unclear since labels are not used to guide the learning of feature. In more complex learning scenarios with an abundance of possible features, these networks could very well learn few label-relevant features, thereby compromising the performance of the top classifier. 

Instead of layer-wise pre-training, several recent approaches train the whole network using a hybrid objective that contains supervised and unsupervised error terms~\cite{Zhao_etal15}. In some of these network configurations, the unsupervised error terms are local to each layer~\cite{Zhang_etal16}. The supervised error term, however, requires backpropagating errors through the whole network. This requirement is avoided in the training approach in~\cite{Ranzato_Szummer08} used to learn to extract compact feature vectors from documents: training proceeds layer by layer where the error in each layer is a combination of a reconstruction error and a supervised error coming from a local classifier. The local auxiliary decoder and classifier pathways still require training, however. Other approaches also make use of a combination of supervised (label-dependent) and unsupervised error signals to train Boltzmann machines as discriminative models~\cite{Larochelle_Bengio08,Goodfellow_etal13a}. Learning in~\cite{Goodfellow_etal13a}, however, is more computationally demanding than our approach as as it involves several iterations to approach the mean-field equilibrium point of the network, and errors are still backpropagated through multiple layers. In~\cite{Larochelle_Bengio08}, multi-layer networks are not considered and only a single layer RBM is used. 

In refs~\cite{Lillicrap_etal16,Nokland_etal16,Baldi_etal16,Neftci_etal17a}, the backpropagation scheme is modified to use random fixed weights in the backward path. This relaxes one of the biologically unrealistic requirements of backpropagation which is weight symmetry between the forward and backward pathways. Errors are still non-local, however, as they are generated by the top layer. A learning mechanism that is able to generate error signals locally is the synthetic gradients mechanism~\cite{Jaderberg_etal16,Czarnecki_etal17} in which errors are generated by dedicated error modules in each layer based only on the layer's activity and the label. The parameters of these dedicated error modules are themselves updated based on errors arriving from higher layers in order to make the error modules better predictors of the true, globally-derived, error signal. Our approach generates errors in a different manner through the use of a local classifier, and each layer receives no error information from the layer above.

\section{Methods}
\label{sec:model}
\begin{figure}[t]
\centering
\includegraphics[width=0.5\textwidth]{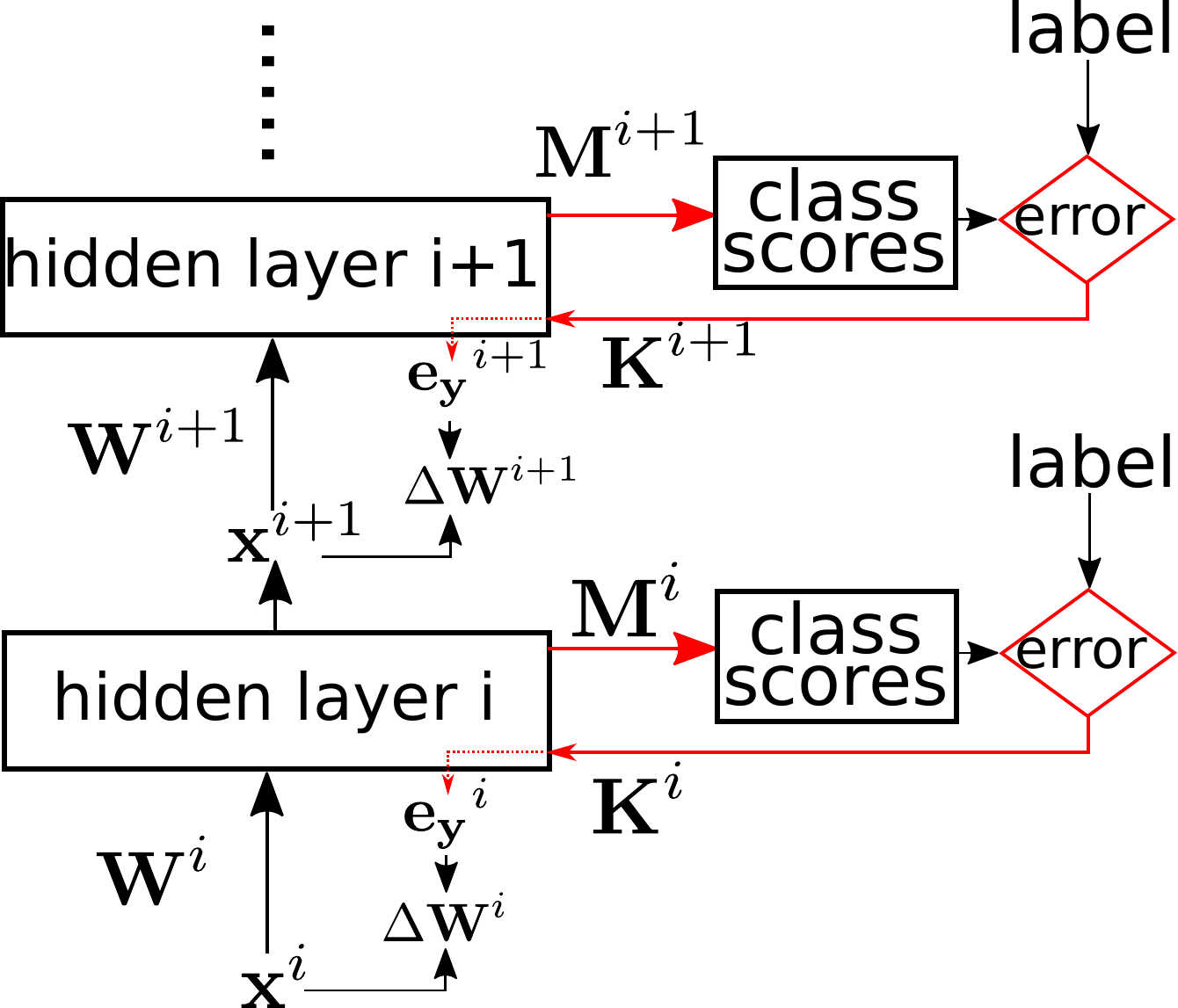}
\caption{Supervised learning in a multi-layer network using local errors. Biases are omitted for clarity. Red arrows indicate the error pathways.  Hidden layer $i$ is trained using local errors generated by a classifier with random fixed weights ${\bf M}^i$. The errors are randomly projected back using the matrix ${\bf K}^i$, and multiplied element-wise with the layer's activation derivative to yield the error signal ${\bf e_y}^i$ which is then used to update the weights.} 
\label{fig:local}
\end{figure}

We train multi-layer networks, with either convolutional or fully connected layers, based on local errors generated by random classifiers. Consider a fully connected $i^{th}$ hidden layer in a network whose activation vector is denoted by ${\bf y}^i\in R^N$ receiving an input ${\bf x}^i \in R^M$: 
\begin{equation}
  {\bf y}^i = f({\bf W}^i{\bf x}^i + {\bf b}^i)
\end{equation}
 \begin{equation}
 {\bf y}{^i}' = f'({\bf W}^i{\bf x}^i + {\bf b}^i)
 \end{equation}
where ${\bf W}^i$ is the $N \times M$ weight matrix of layer $i$ and ${\bf b}^i \in R^N$ is the bias vector, and $f$ 
is the neuron's activation function. 
In all the networks we train, we use Rectified Linear Units (ReLUs)~\cite{Nair_Hinton10}, i.e, $f(x)=max(x,0)$, with corresponding derivatives $f'(x)=H(x)$ where $H(\cdot)$ is the Heaviside step function. We pre-define for this hidden layer a fixed random classifier matrix ${\bf M}^i$ which is a $C \times N$ matrix where $C$ is the number of classification categories. The random matrix, ${\bf M}^i$, is used to convert the layer activation vector, ${\bf y}^i$, to a category score vector ${\bf s}^i \in R^C$ where ${\bf s}^i = {\bf M}^i{\bf y}^i$. Since this is a supervised learning setting, the correct input category $t$ is known during training, which allows the layer to generate a scalar loss or error signal, $E(t,{\bf s}^i)$. $E$ could be for example the cross-entropy loss or the square hinge loss. This error is then backpropagated in order to calculate the weight and bias updates, $\Delta {\bf W}^i$ and $\Delta {\bf b}^i$:
\begin{align}
  {\bf e_s}^i &= \frac{dE}{d{\bf s}^i} \\
  {\bf e_y}^i &= {\bf K^i}{\bf e_s^i} \odot {\bf y}{^i}'  \\
  \Delta{\bf W}^i &= -\eta {\bf e_y}^i {\times} {\bf x}^i \\
  \Delta{\bf b}^i &= -\eta {\bf e_y}^i
\end{align}
where $\odot$ is the element-wise multiplication operator, $\times$ is the outer product operator, and $\eta$ is the learning rate. ${\bf K}^i$ is the $N \times C$ matrix used to backpropagate the classifier error to the layer being trained. If we set ${\bf K}^i = {\bf M}{^i}^T$, then the weight and bias updates are executing exact gradient descent to minimize the random classifier error, $E$. In that case, training of each layer is equivalent to training a network with one hidden layer where only the hidden layer's input weights and biases are trainable, while the output weights, ${\bf M}^i$ are fixed. The learning scheme is illustrated in Fig.~\ref{fig:local}. For convolutional layers, the learning scheme remains unchanged. The post-activation feature maps tensor is simply flattened to yield a 1D vector before multiplying by the random classifier matrix ${\bf M}$.

We also use dropout~\cite{Srivastava_etal14} in this training setting to minimize overfitting. All incoming/outgoing weights to/from a dropped neurons are not updated in the iteration in which the neuron is dropped.  In some networks, we use batch normalization~\cite{Ioffe_Szegedy15} before the layer's non-linearity. The layer's learnable parameters will then include a scaling factor (one for each neuron in a fully connected layer, or one for each feature map in a convolutional layer) that is also trained using local errors. For a fully connected layer, the input to the local classifier is taken after the dropout mask is applied (if dropout is used). For a convolutional layer, the input to the layer's local classifier is taken after pooling and after applying the dropout mask.  In all experiments, we initialize the fixed random classifier weights, as well as the trainable weights in the main network, from a uniform, zero-mean, distribution whose max/min values depend on the number of neurons in the source and target layers according to the scheme in~\cite{Glorot_Bengio10}.  

We compare our approach to the feedback alignment training method~\cite{Lillicrap_etal16} in which random fixed weights are used to backpropagate the error layer-by-layer from the top layer. The layer's activation derivative is still used when backpropagating errors. In the presence of max-pooling layers, errors only backpropagate through the winner(max) neuron in each pooling window. When using feedback alignment training in the presence of dropout, a neuron that is dropped during the forward pass is also dropped during the backward pass. When using convolutional layers, we use fixed random filters that we convolve with the errors of one convolutional layer to yield the errors at the outputs of the previous/lower convolutional layer. We also use batch normalization when training using feedback alignment. The extra scaling parameters introduced by batch normalization are trained using the randomly backpropagated errors arriving at the batch-normalized layer's output. 

All experiments in this paper were carried out using Theano~\cite{Bastien_etal12,Bergstra_etal10}, and all parameters were optimized using ADAM~\cite{Kingma_Ba14}.

\section{Hardware Implications of Learning Using Local Errors}
\label{sec:hw}
\begin{figure}[t]
\centering
  \centering
  \begin{subfigure}[b]{0.62\textwidth}
    \includegraphics[width=\textwidth]{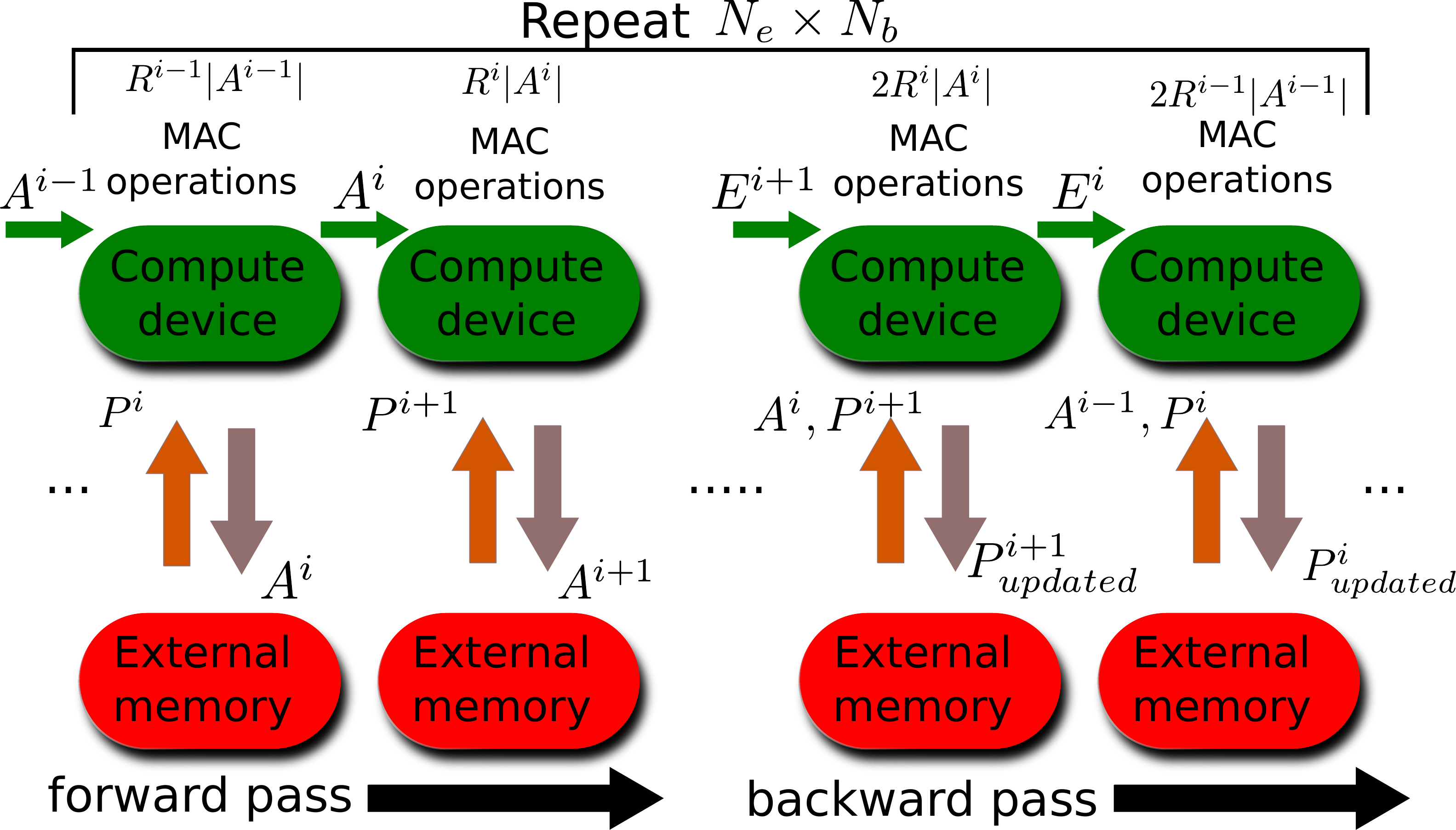}
    \subcaption{}
    \label{fig:traffic_a}
  \end{subfigure}
  \quad
  \begin{subfigure}[b]{0.42\textwidth}
    \includegraphics[width=\textwidth]{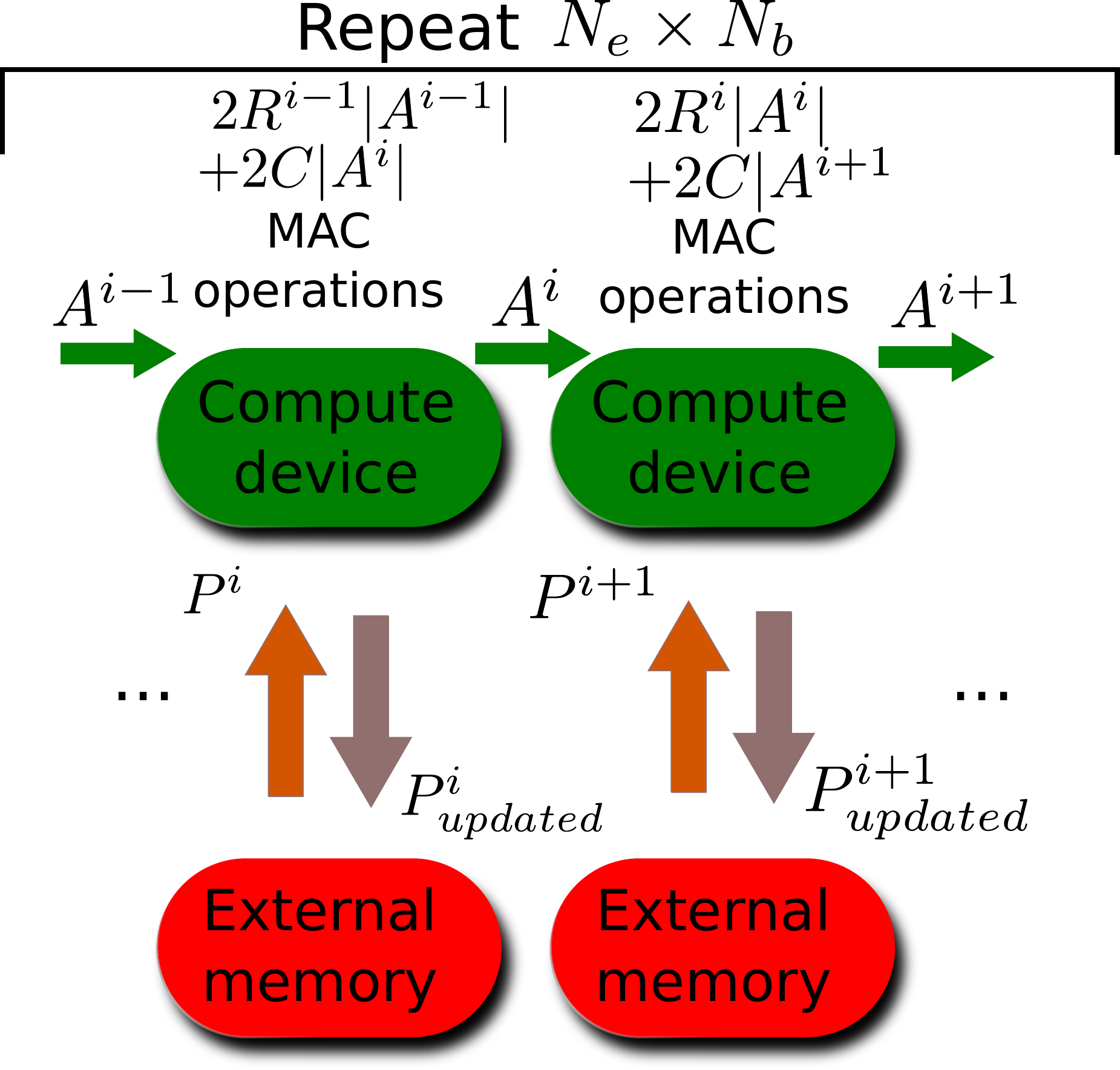}
    \subcaption{}
    \label{fig:traffic_b}
  \end{subfigure}
\caption{Memory traffic and number of MAC operations for different training methods. Arrows between compute device and external memory indicate memory traffic while green arrows indicate data buffered and reused by the compute device. Each computation stage is executed a number of times given by the enclosing repeat block. (\subref{fig:traffic_a}) Standard backpropagation learning.  (\subref{fig:traffic_b}) Training all layers simultaneously using local errors. Note that there is no backward pass as weights are updated during the forward pass.} 
\label{fig:traffic}
\end{figure}

Standard learning techniques based on backpropagating errors through the whole network require the hardware executing the learning algorithm to store the activation values and activation derivatives of all network layers in order to calculate weight updates and backpropagate errors once errors are available from the top layer. This imposes several communication and memory access overheads if learning is executed on hardware whose memory can not accommodate all the network weights and activations. For large scale convolutional networks, this practically includes all CPU and GPU devices where on-chip memory is limited to few tens of MBytes, while state of the art deep convolutional networks typically require several hundred MBytes to several GBytes in order to store the network weights and mini-batch activations~\cite{Rhu_etal16}. Data thus has to be continuously shuttled between the compute device and external memory. This is the case even in custom accelerators developed to accelerate just the inference (feed-forward) phase~\cite{Himavathi_etal07,Cavigelli_etal15,Chen_etal16,Han_etal16,Ardakani_etal16,Aimar_etal17,Jouppi_etal17}, where a complete forward pass through a large-scale convolutional network can not be executed completely on the accelerator without having to access external memory to store intermediate activations and to load weights.

Improvements in memory bandwidth significantly lag improvements in computing elements speed~\cite{Wulf_McKee95}. Reducing memory traffic in a compute intensive task such as learning deep networks thus improves performance as it relaxes the requirements on the memory bandwidth and latency needed to keep the compute elements occupied, allowing either the compute elements to run at higher frequencies or the external memory to run at lower frequencies. Moreover, energy needed to drive off-chip traffic from/to external memory as well as memory read/write energy often contribute significantly to the overall energy consumption~\cite{Vogelsang10,Lefurgy_etal03}. Reducing memory traffic can thus have significant impact on the overall energy consumption of the learning hardware. In this section, we analyze the savings in memory traffic volume obtained using the learning approach based on local errors that we propose in this paper. 

Consider a neural network with $L$ layers. $P^i$ and $A^i$ are the parameters and the mini-batch activations of layer $i$, respectively. $\lvert P^i\rvert$ and $\lvert A^i\rvert$ are the number of elements in $P^i$ and $A^i$. A neuron in layer $i$ has a fanout of $R^i$, i.e, a neuron in layer $i$ projects to $R^i$ neurons in layer $i+1$. In convolutional layers, we ignore any border effects which might cause the neurons at the borders of the feature maps to project to fewer neurons than neurons away from the borders. We divide the training data set into $N_{b}$ mini-batches and train the network for $N_{e}$ epochs. Each weight and each neuron activation takes up one memory word (which we assume is 32 bits).

Figure~\ref{fig:traffic_a} illustrates the data traffic and the number of MAC operations needed during standard backpropagation training. The data traffic in Fig.~\ref{fig:traffic_a} assumes the compute device has enough on-board memory to buffer the output activations of one layer in order to use these activations to calculate the next layer's activation. We also assume the compute device does not need the parameters of any layer to be streamed in more than once during each forward pass and during each backward pass. These assumptions would hold true if the accelerator has at least  $\max_i (\lvert P^i\rvert +  \lvert A^i\rvert)$ words of on-board memory. During the forward pass, the activations of all layers have to be streamed out to external memory so they can be used in the backward pass. The number of MAC operations needed to calculate the activation of layer $i$ is $R^{i-1}\lvert A^{i-1} \rvert$. During the backward pass, the compute device buffers the back-propagated errors of one layer and uses them to calculate the errors at the preceding layer. $R^{i}\lvert A^{i} \rvert$ MAC operations are needed  to calculate the weight updates for $P^{i+1}$. An additional $R^{i}\lvert A^{i} \rvert$ MAC operations are needed to backpropagate the errors from layer $i+1$ to layer $i$. We ignore the special case of the input layer where errors do not need to be backpropagated. We also ignore the MAC operations needed to calculate the error at the top layer. 

Figure~\ref{fig:traffic_b} illustrates the case when learning is done using errors generated by random local classifiers. As in standard backpropagation, $R^{i-1}\lvert A^{i-1} \rvert$ MAC operations are needed to calculate the activations of layer $i$. To calculate the local classifier output, $C \lvert A^i \rvert$ MAC operations are needed where $C$ is the number of classification classes. Note that the random classifier weights can be generated on the fly using a PRNG, and thus only require the PRNG seed (whose size can be 32 bits for 32-bit weights) to be stored. To backpropagate the local classifier error to obtain the error at layer $i$, an additional $C \lvert A^i \rvert$ MAC operations are needed and $R^{i-1}\lvert A^{i-1} \rvert$ MAC operations are needed to update the parameters of layer $i$, $P^i$, based on the layer's error. 

Table~\ref{table:traffic_macs} summarizes the number of MAC operations and the memory read/write volume required by the two training methods. Learning using local errors has a decisive advantage when it comes to memory traffic as it requires drastically less read and write operations compared to standard backpropagation. The reduction in the number of MAC operations is less unequivocal as it depends on the number of classification classes, $C$, and the fanout of the neurons in the network, $R^i$. Learning using local error reduces the MAC operations count if $L\times C < 0.5\sum_i  R^i$. This condition is easily satisfied when the number of classes is small and it was satisfied by all the networks presented in this paper.

\begin{table}[h]
  \caption{Memory traffic and number of MAC operations for different learning methods}
  \label{table:traffic_macs}
  \centering
  \begin{tabular}[t]{lllp{35mm}}
    \toprule
    Training method     & Memory read(words)  & Memory write (words) & MAC operations    \\
    \midrule
    Standard backpropagation  & $N_{e}N_{b} \sum\limits_i (2\lvert P^i \rvert  + \lvert A^i \rvert )$   & $N_{e}N_{b} \sum\limits_i (\lvert P^i \rvert  + \lvert A^i \rvert)$ & $N_{e}N_{b} \sum\limits_i 3R^{i}\lvert A^i \rvert$     \\
    Learning using local errors     & $N_e N_b \sum\limits_i \lvert P^i \rvert $ & $N_e N_b \sum\limits_i \lvert P^i \rvert$    & $N_eN_b\sum\limits_i(2R^i + 2C)\lvert A^i \rvert$   \\

    \bottomrule
  \end{tabular}
\end{table}

\section{Results}
\label{sec:results}
\subsection{MNIST}
\begin{figure}[h]
\centering
\begin{subfigure}[b]{0.48\textwidth}
  \includegraphics[width=\textwidth]{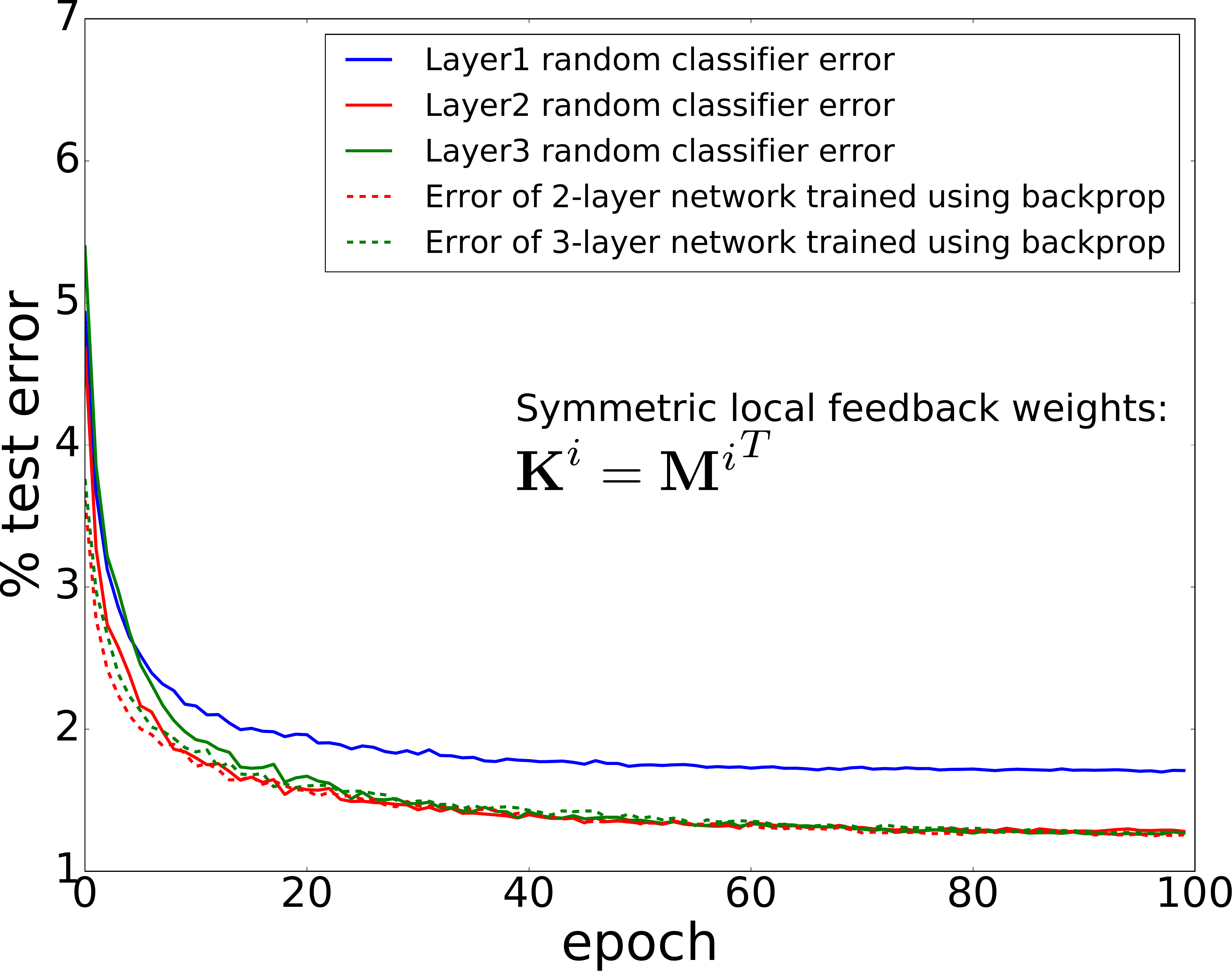}
  \subcaption{}
  \label{fig:mnist_symmetric}
\end{subfigure}
\quad
\begin{subfigure}[b]{0.48\textwidth}
  \includegraphics[width=\textwidth]{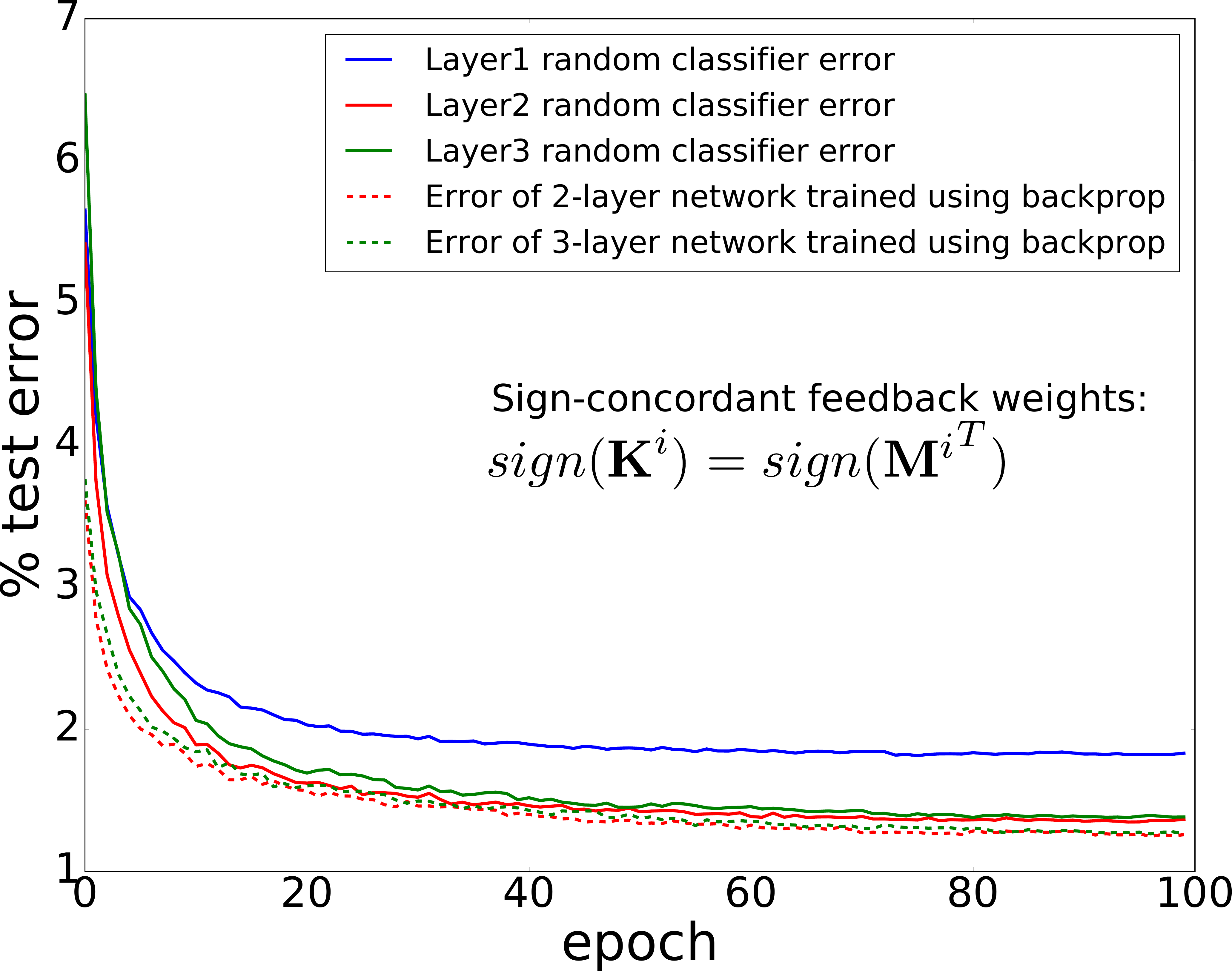}
  \subcaption{}
  \label{fig:mnist_concordant}
\end{subfigure}

    \caption{(\subref{fig:mnist_symmetric}) MNIST test set errors obtained from three networks: a 3-layer network trained using local errors and symmetric local feedback weights (${\bf K}^i = {\bf M}{^i}^T$ ) where the errors for the three random local classifiers are shown, a network with two hidden layers trained using standard backpropagation, and a network with three hidden layers trained using standard backpropagation. The networks were trained for 100 epochs. Each line is the average of 20 training trials. (\subref{fig:mnist_concordant}) Same as (\subref{fig:mnist_symmetric}) except that in the network trained using local errors, sign-concordant local feedback weights with independent and random magnitudes were used. }
    \label{fig:mnist_results}
\end{figure}

We first validate the performance of our training approach on the MNIST hand-written digit recognition task. We used the standard split of 50,000/10,000/10,000 examples for training/validation/testing respectively. The validation set was added to the training set after choosing the hyper-parameters. We use a network with 3 fully connected hidden layers with $1000$ neurons per layer and train the weights in the entire network using local errors. As a baseline, we also train a 2-hidden layers network and a 3-hidden layers network using standard backpropagation where each hidden layer also has $1000$ neurons. Dropout was used in all networks to reduce overfitting. We first used fixed symmetric random weights in the forward and backward pathways in the local error loops, i.e, ${\bf K}^i = {\bf M}{^i}^T$ in all layers. The results are shown in Fig.~\ref{fig:mnist_symmetric}. The local classifier errors improve for the second and third hidden layers compared to the first hidden layer implying that the network is able to make use of depth to obtain better accuracy. The local classifier errors in the second and third layers are similar implying that the network is unable to make use of the increased depth beyond two hidden layers for this simple dataset. This observation is also valid for standard backpropagation where accuracy does not improve when going from two hidden layers to three hidden layers. When training using local errors, we also ran experiments where the local classifier weights were trainable parameters. This had no effect on accuracy as shown in Table~\ref{tab:mnist}.

\begin{table}[h]
  \caption{MNIST final test set error after 100 training epochs. When learning using local errors, the local classifier errors in all layers are reported. Mean and standard deviation from 20 runs.}
  \centering
  \begin{tabular}[t]{lp{35mm}p{35mm}p{35mm}}
    \toprule
         & Local error learning (symmetric feedback weights) & Local error learning (sign-concordant feedback weights) & Local error learning (trainable local classifier)     \\
    \midrule
    Test error  & \parbox{5cm}{Layer1: $1.71 \pm 0.042 \%$   \\ Layer2: $1.28 \pm 0.042 \%$ \\   Layer3: $1.27 \pm 0.048 \%$} &   \parbox{5cm}{Layer1: $1.83 \pm 0.048 \%$   \\ Layer2: $1.37 \pm 0.055 \%$ \\   Layer3: $1.38 \pm 0.059 \%$} &  \parbox{5cm}{Layer1: $1.42 \pm 0.039 \%$   \\ Layer2: $1.28 \pm 0.049 \%$ \\   Layer3: $1.27 \pm 0.037 \%$}   \\
    \bottomrule \mbox{} \\ \toprule
    & Learning using feedback alignment & 2-layer network trained using backprop & 3-layer network trained using backprop \\
    \midrule
    Test error  &  $1.70 \pm 0.087 \%$ & $1.26 \pm 0.068 \%$  & $1.27 \pm 0.050 \%$ \\
    \bottomrule
  \end{tabular}
  \label{tab:mnist}

\end{table}

Next, to lessen concern of biological implausibility of exact symmetry in feedforward and feedback weights, we relaxed the weight symmetry requirement in the local error loops and initialized the error feedback weights ${\bf K}^i$ randomly and independently of ${\bf M}{^i}$, except we then modified the sign of the weights in ${\bf K}^i$ so that $sign({\bf K}^i) = sign({\bf M}{^i}^T)$. The signs of the feedback weights in the local error loops thus match the signs of the feedforward weights (both are fixed and have independent magnitudes). This is the 'sign-concordant feedback weights' case shown in Fig.~\ref{fig:mnist_concordant}. Performance deteriorates slightly in this case compared to symmetric feedforward and feedback local classifier weights. When we relax the symmetry requirement further and choose ${\bf K}^i$ to be random and completely independent of ${\bf M}{^i}$, the network failed to learn and error rates stayed at near-chance level. We also experimented with training based on feedback alignment where errors from the top layer are backpropagated using random fixed weights. The network's performance using feedback alignment is worse than learning using local errors (using either symmetric or sign-concordant weights) as shown in Table~\ref{tab:mnist}. 

It is important to note that in feedback alignment, the feedforward weights eventually 'align' with the random weights used to backpropagate errors~\cite{Lillicrap_etal16} enabling the network to learn. When learning using random fixed local classifiers, and if we choose random error feedback weights, the classifier weights are fixed and thus can not align with the random weights used in the one-step backpropagation. Reliable error information, however, can still reach the layer being trained if the signs of the random backpropagation weights, ${\bf K}^i$, match the signs of the fixed local classifier weights ${\bf M}^i$. This is in-line with previous investigations into the importance of weight symmetry in backpropagation that argue for the importance of sign-concordance between forward and backward weights~\cite{Liao_etal16}.

\subsection{CIFAR10}
We trained a convolutional network with three convolutional layers followed by two fully connected layers on the CIFAR10 dataset. We used a similar network as ref.~\cite{Srivastava_etal14}. The convolutional layers used a $5\times 5$ kernel, a stride of $1$, and had $96$, $128$, and $256$ feature maps going from the bottom upwards. Max-pooling with a pooling window of $3 \times 3$ and stride $2$ was applied after each convolutional layer. The two fully connected layers on top had $2,048$ neurons each. All layers were batch-normalized and dropout was applied after the input layer, after each max-pooling layer, and after each fully connected layer.

\begin{figure}[h]
\centering
\begin{subfigure}[b]{0.48\textwidth}
  \includegraphics[width=\textwidth]{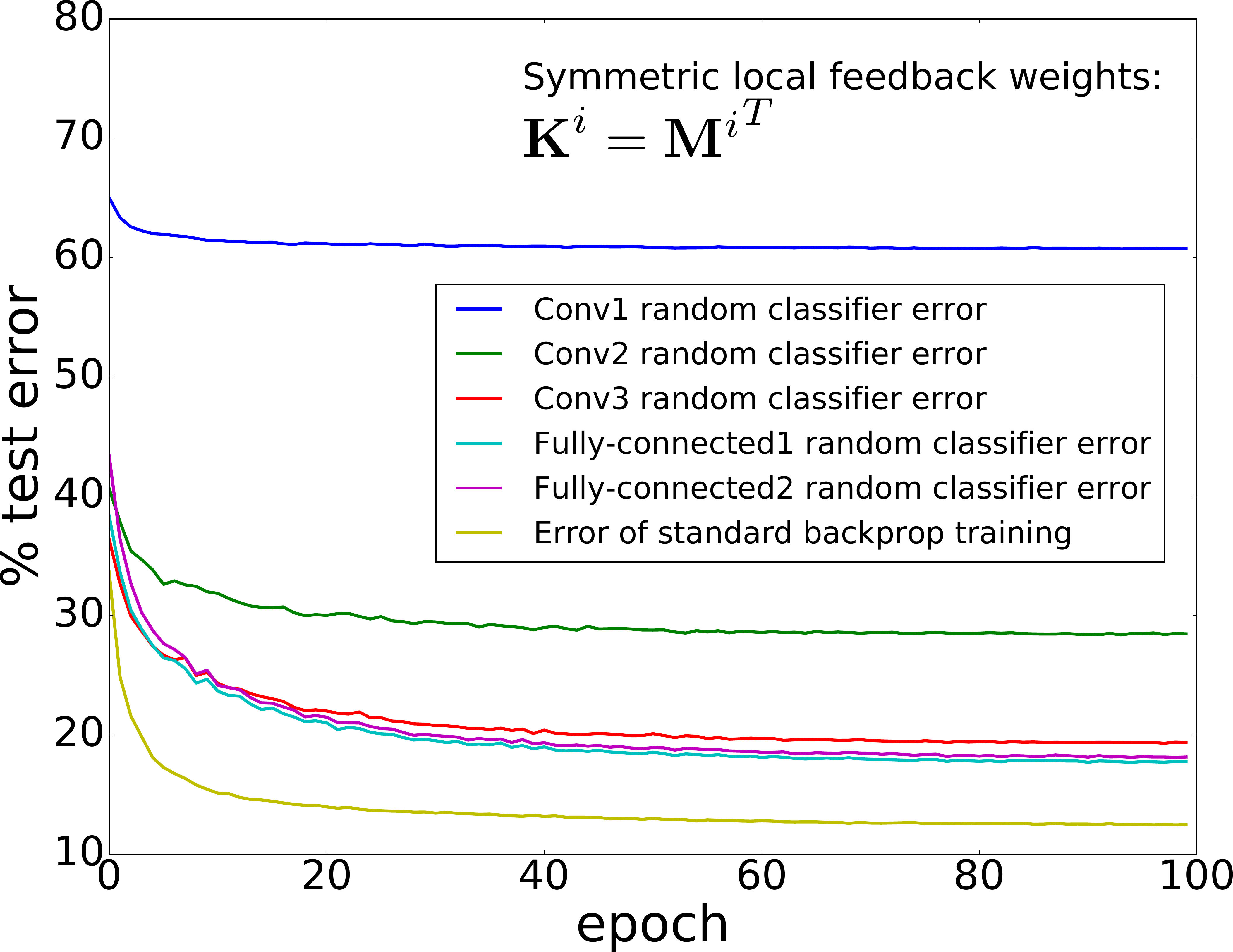}
  \subcaption{}
  \label{fig:cifar10_symmetric}
\end{subfigure}
\quad
\begin{subfigure}[b]{0.48\textwidth}
  \includegraphics[width=\textwidth]{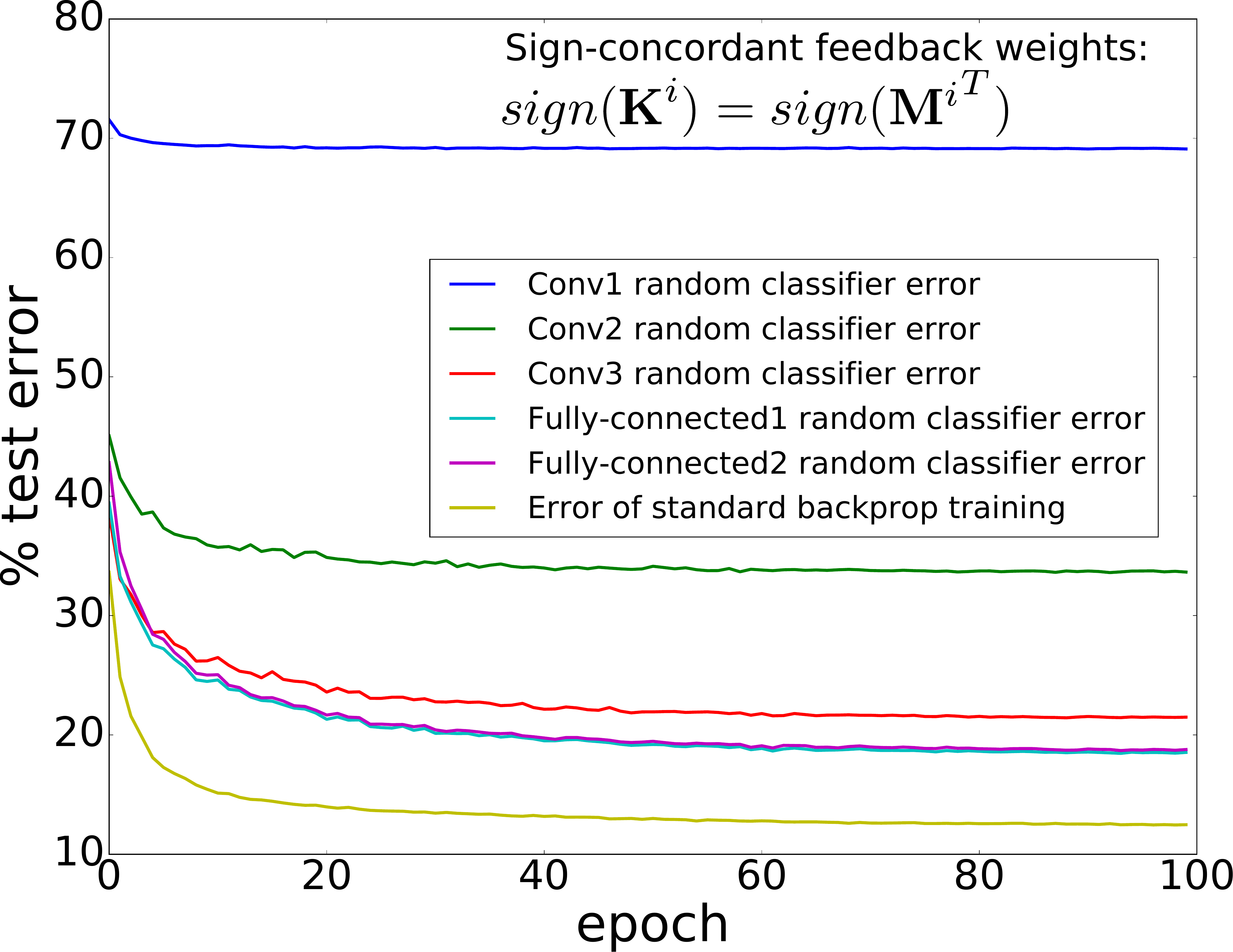}
  \subcaption{}
  \label{fig:cifar10_concordant}
\end{subfigure}

    \caption{(\subref{fig:cifar10_symmetric}) CIFAR10 test set errors obtained from two convolutional networks having the architecture described in the main text: one network was trained using local errors and symmetric local feedback weights (${\bf K}^i = {\bf M}{^i}^T$ ), where the errors for the random local classifiers in all layers are shown; the other network with identical architecture was trained using standard backpropagation. (\subref{fig:cifar10_concordant}) Same as (\subref{fig:cifar10_symmetric}) except that in the network trained using local errors, sign-concordant local feedback weights with independent and random magnitudes were used. }
    \label{fig:cifar10_results}
\end{figure}

The 32$\times$32$\times$3 CIFAR10 color images were pre-processed using global contrast normalization followed by ZCA whitening. The training set of 50,000 images was used for training/validation and we report errors on the 10,000 images test set. Unlike the MNIST dataset, standard backpropagation significantly outperforms training using local errors as shown in Fig.~\ref{fig:cifar10_results} and Table~\ref{tab:cifar10}. Performance of local error learning deteriorates slightly when using sign-concordant local feedback weights instead of symmetric local feedback weights. Performance does not improve for the local classifier in the second fully connected layer compared to the classifier in the first fully connected layer. We trained a variant of the network using only one fully connected layer using standard backpropagation. As shown in Table~\ref{tab:cifar10}, the improvement in performance of the network trained using standard backpropagation is minimal when going from one to two fully connected layers. This implies that the second fully connected layer is largely superfluous and local error learning is thus unable to capitalize on it. Unlike for the MNIST dataset, allowing the local classifier parameters to be trainable improves performance significantly. As was the case for the MNIST dataset, training using feedback alignment leads to significantly worse performance than learning using local errors.

\begin{table}[h]
  \caption{CIFAR10 final test set error after 100 training epochs. When learning using local errors, the local classifier errors in all layers are reported. Mean and standard deviation from 20 runs.}
  \centering
  \begin{tabular}[t]{lp{40mm}p{40mm}p{40mm}}
    \toprule
         & Local error learning (symmetric feedback weights) & Local error learning (sign-concordant feedback weights) & Local error learning (trainable local classifier)     \\
    \midrule
    Test error  & \parbox{5cm}{conv1: $60.74  \pm 0.87 \%$ \\ conv2: $28.47 \pm 0.37 \%$ \\ conv3: $19.37 \pm 0.27 \%$ \\ FC1: $17.75 \pm 0.30 \%$ \\ FC2:  $18.15 \pm 0.36 \%$} &   \parbox{5cm}{conv1: $69.10  \pm 1.1 \%$ \\ conv2: $33.64 \pm 0.41 \%$ \\ conv3: $21.49 \pm 0.38 \%$ \\ FC1: $18.5 \pm 0.33 \%$ \\ FC2:  $18.77 \pm 0.33 \%$}  &  \parbox{5cm}{conv1: $30.45  \pm 0.27 \%$ \\ conv2: $18.44 \pm 0.27 \%$ \\ conv3: $15.07 \pm 0.21 \%$ \\ FC1: $14.34 \pm 0.24 \%$ \\ FC2:  $14.34 \pm 0.22 \%$}   \\
    \bottomrule \mbox{} \\ \toprule
    & Learning using feedback alignment & Learning using standard backpropagation (two FC layers) & Learning using standard backpropagation (one FC layer)  \\
    \midrule
    Test error  &  $20.87 \pm 0.34 \%$ & $12.47 \pm 0.25 \%$  & $12.72 \pm 0.21 \%$  \\
    \bottomrule
  \end{tabular}
  \label{tab:cifar10}

\end{table}

Our feedback alignment results are better than those previously reported in refs.~\cite{Baldi_etal16,Liao_etal16,Nokland_etal16}. This is due to our use of a bigger network that is well-regularized using dropout. Using a well-regularized network is particularly crucial when investigating alternatives to standard backpropagation as poorly-regularized learning can make a worse learning algorithm seem better, simply because it better regularizes the learning problem compared to a superior algorithm that overfits on the training data. Strong regularization is also a potential reason why we see that exact gradient descent (standard backpropagation) is clearly superior, unlike previous investigations that report better performance when using various approximations to standard backpropagation~\cite{Liao_etal16}, where this better performance can be due to the better regularization introduced by the approximate learning algorithms.

\subsection{SVHN}
We trained an identical network to the one used for the CIFAR10 dataset on the SVHN dataset. The SVHN dataset is a dataset of 32$\times$32$\times$3 color images. We used the training/validation/testing split of 598388/6000/26032 images respectively that was previously used in refs.~\cite{Srivastava_etal14,Goodfellow_etal13,Sermanet_etal12}. The validation set was added to the training set after choosing the hyper-parameters (learning rate and dropout probabilities). The images were preprocessed using the local contrast normalization technique from ref.~\cite{Jarrett_etal09}.

\begin{figure}[h]
\centering
\begin{subfigure}[b]{0.48\textwidth}
  \includegraphics[width=\textwidth]{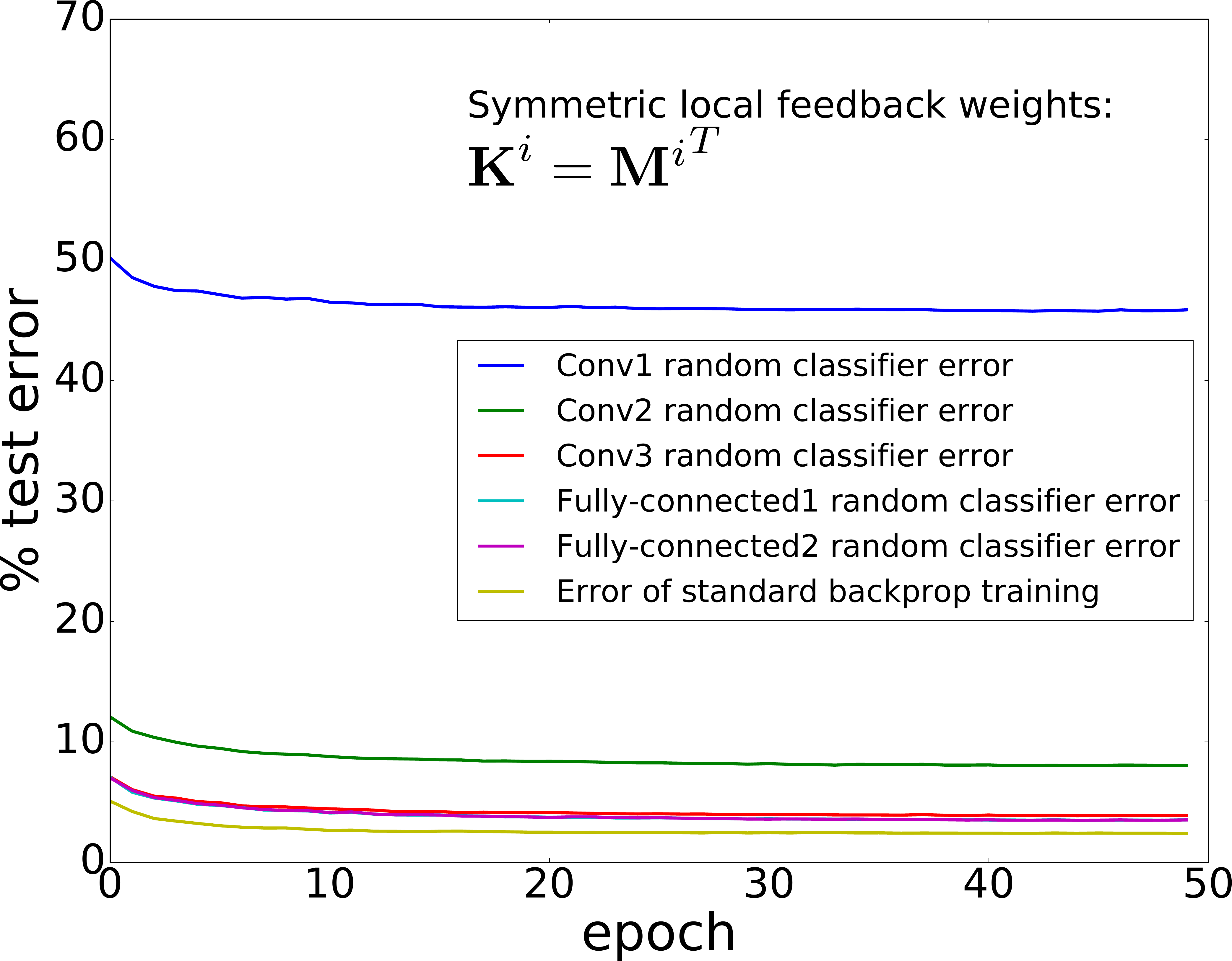}
  \subcaption{}
  \label{fig:svhn_symmetric}
\end{subfigure}
\quad
\begin{subfigure}[b]{0.48\textwidth}
  \includegraphics[width=\textwidth]{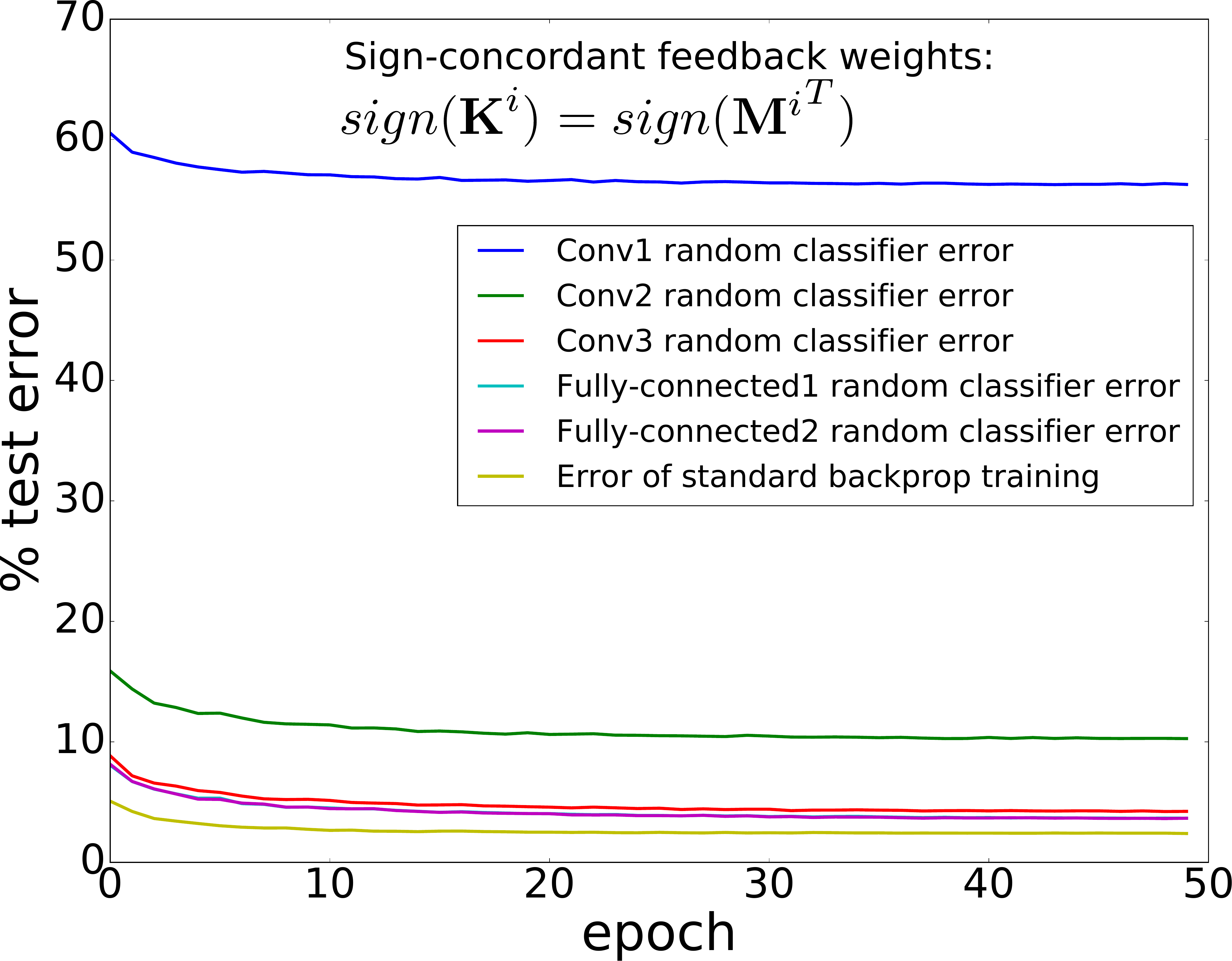}
  \subcaption{}
  \label{fig:svhn_concordant}
\end{subfigure}

    \caption{(\subref{fig:svhn_symmetric}) SVHN test set errors obtained from two convolutional networks: one network was trained using local errors and symmetric local feedback weights (${\bf K}^i = {{\bf M}^i}^T$) where the errors for the random local classifiers in all layers are shown. The other network with identical architecture was trained using standard backpropagation . (\subref{fig:cifar10_concordant}) Same as (\subref{fig:cifar10_symmetric}) except that in the network trained using local errors, sign-concordant local feedback weights with independent and random magnitudes were used. }
    \label{fig:svhn_results}
\end{figure}

Figure~\ref{fig:svhn_results} shows the test error curves for the case of the symmetric local feedback weights and the case of the sign-concordant local feedback weights. The test error trends in Fig.~\ref{fig:svhn_results} and Table~\ref{tab:svhn} are similar to those observed for CIFAR10. The performance of standard backpropagation is clearly superior, followed by learning using local errors generated by trainable local classifiers. Learning using local errors generated by fixed random classifiers lags behind (both when using symmetric feedback weights or sign-concordant feedback weights) but it still outperforms learning using feedback alignment.

\begin{table}[h]
  \caption{SVHN final test set error after 100 training epochs. When learning using local errors, the local classifier errors in all layers are reported. Mean and standard deviation from 5 runs.}
  \centering
  \begin{tabular}[t]{lp{40mm}p{40mm}p{40mm}}
    \toprule
         & Local error learning (symmetric feedback weights) & Local error learning (sign-concordant feedback weights) & Local error learning (trainable local classifier)     \\
    \midrule
    Test error  & \parbox{5cm}{conv1: $45.87  \pm 0.64 \%$ \\ conv2: $8.05 \pm 0.19 \%$ \\ conv3: $3.87 \pm 0.037 \%$ \\ FC1: $3.53 \pm 0.022 \%$ \\ FC2:  $3.52 \pm 0.032 \%$} &   \parbox{5cm}{conv1: $56.27  \pm 0.86 \%$ \\ conv2: $10.27 \pm 0.28 \%$ \\ conv3: $4.23 \pm 0.086 \%$ \\ FC1: $3.66 \pm 0.091 \%$ \\ FC2:  $3.66 \pm 0.084 \%$}  &  \parbox{5cm}{conv1: $9.96  \pm 0.062 \%$ \\ conv2: $3.83 \pm 0.10 \%$ \\ conv3: $2.79 \pm 0.066 \%$ \\ FC1: $2.57 \pm 0.049 \%$ \\ FC2:  $2.57 \pm 0.019 \%$}   \\
    \bottomrule \mbox{} \\ \toprule
    & Learning using feedback alignment & Learning using standard backpropagation &  \\
    \midrule
    Test error  &  $3.74 \pm 0.077 \%$ & $2.39 \pm 0.037 \%$  &  \\
    \bottomrule
  \end{tabular}
  \label{tab:svhn}

\end{table}

\section{Conclusions and Discussion}
\label{sec:conclusions_discussion}
Weight symmetry between the forward and backward passes and delayed error generation are two of the most biologically unrealistic aspects of backpropagation. Recent investigations have shown that the weight symmetry requirement can be relaxed allowing learning to proceed with random feedback weights~\cite{Lillicrap_etal16,Baldi_etal16,Nokland_etal16,Neftci_etal17a}. These investigations, however, do not address the problem of local learning and require the network to maintain its state until errors arrive from higher layers. Local errors have often been used to augment the top layer errors~\cite{Szegedy_etal14,Lee_etal15}. However, until now, relatively little work has been done on supervised learning using exclusively local errors, and none that we know of investigated local error generation using fixed random classifiers.

Our results show that learning using local errors generated using random classifiers, while falling short of the performance of standard backpropagation, significantly outperforms learning using feedback alignment techniques~\cite{Lillicrap_etal16,Baldi_etal16}. This holds true even when relaxing the weight symmetry requirement in the local feedback loop and using random fixed feedback weights that are sign-aligned with the random fixed classifier weights in the local learning loop. Maintaining sign-alignment is problematic in the feedback alignment technique as the sign of the feedback weights have to dynamically track the sign of the feedforward weights during training~\cite{Liao_etal16} which introduces a dynamic dependency between the two sets of weights. In our case, since both sets of weights are fixed, this dependency need only be enforced initially.

Our CIFAR10 and SVHN results indicate that locally generated errors allow a convolutional layer to learn good features that are then used by the subsequent layer to learn even more informative features as evidenced by the increased accuracy of the local classifiers in higher layers. In the end, however, our approach solves many small optimization problems where each problem involves only the weights of one layer. We therefore lose one of the core advantages of standard backpropagation learning using a global objective function: the high probability of finding a good minimum in the parameter space when the dimensionality of this parameter space is large, i.e, when it includes all the network parameters~\cite{Choromanska_etal15,Im_etal16}. It was thus expected that classification performance will suffer compared to learning using standard backpropagation and a global objective function.

Single cell measurements in monkey area IT indicate broad tuning to a range of categories~\cite{Kiani_etal07,Sigala_Logothetis02}. This broad category tuning is realized in the proposed training scheme through the random local classifier weights that define how a neuron contributes to the score of each classification category. During training, the actual tuning properties of each neuron change to be in-line with the pre-defined fixed tuning defined by the random classifier weights, as this is the only way to minimize the local classifier error. Our error generation mechanism has several biologically attractive aspects:
\begin{enumerate}
\item It involves only two synaptic projections allowing errors to be generated quickly and weight updates to be carried out before input-induced changes in the states of the neurons have decayed. This avoids the common and unrealistic input buffering requirement encountered in standard backpropagation and feedback alignment techniques.
\item Error generation involves random projections that do not have to be learned. This makes the error generation loop particularly simple and removes any potential problematic interactions between learning the auxiliary classifier weights and learning the main network weights.
\item Strict weight symmetry is not required in the error pathway, only sign-alignment between two sets of fixed random weights is needed. 
\end{enumerate}

The use of fixed random local classifier weights allows us to sidestep one of the main hardware-related issues of using auxiliary local classifiers: the need to store the local classifier weights. Especially in large convolutional layers, storing the local classifier weights could be prohibitively expensive in terms of memory resources. Since the local classifier weights need to be accessed in a fixed order during each training iteration in order to calculate the classifier outputs, they can be cheaply, quickly, and reproducibly generated on the fly using a PRNG and a small seed. We have shown that this approach allows us to obtain a learning mechanism that drastically reduces memory traffic compared to standard backpropagation. During inference, the random classifier weights in each layer (which are compactly stored in a small seed) can be used to generate a classification decision during the evaluation of each layer. Thus, if needed, a series of classification decisions can be obtained, one from each layer, at a small computational cost and virtually no memory cost. The decisions from bottom layers, even though less accurate than the decisions from higher layers, can be used in situations where response time is critical. This allows the network to be dynamically truncated where higher layers are not evaluated and the final decision taken from intermediate layers. This feature of the proposed networks enables a dynamical trade-off between accuracy and energy consumption/computational load where only as many layers as allowed by the energy budget, or response time constraint, are evaluated.

\end{document}